\documentclass[runningheads]{llncs}

\usepackage[T1]{fontenc}
\usepackage{graphicx}
\usepackage{amsmath,amssymb}
\usepackage{booktabs}
\usepackage{xcolor}
\usepackage{multirow}

\begin{document}

\title{Inverse Bayesian Inference for Extracting Lesion Dynamics from Longitudinal Spectral CT}
\titlerunning{Inverse Bayesian Inference}

\author{Lukas F\"orner\inst{1,2,3,4} \and 
Melina W\"ordehoff\inst{1,2,4} \and 
Julian Steffens\inst{1} \and 
Maximilian Schmutz\inst{2,5,6} \and 
Rainer Claus\inst{4,5,6,7} \and 
Josua Decker\inst{1} \and 
Thomas Kr\"oncke\inst{1,4} \and 
Kartikay Tehlan\inst{1,2,3,4} \and 
Thomas Wendler\inst{1,2,3,4,8}}
\authorrunning{L. F\"orner et al.}
\institute{Dept. of diagnostic and interventional Radiology and Neuroradiology, University Hospital Augsburg, Germany \\ \textbf{\email{lukas.foerner@uk-augsburg.de}}\and
Digital Medicine, University Hospital Augsburg, Germany \and
Chair for Computer-Aided Medical Procedures and Augmented Reality, Technical University of Munich, Germany
\and
Bavarian Center for Cancer Research (BZKF) Augsburg, Germany \and
Dept. of Hematology and Oncology, University Hospital Augsburg, Germany \and
Comprehensive Cancer Centre Augsburg (CCCA), University Hospital Augsburg, Germany \and
Dept. of General Pathology and Molecular Diagnostics, University Hospital Augsburg, Germany \and
Center for Advanced Analytics and Predictive Sciences, University of Augsburg, Germany\\}

\maketitle

\begin{abstract}
Longitudinal medical imaging captures temporal evolution of lesions, yet extracting the underlying dynamical parameters governing this evolution remains challenging. We propose an inverse Bayesian framework for inferring lesion dynamics from longitudinal spectral CT. We decompose spectral feature ($x$) evolution into three components:
\begin{equation*}
    \frac{dx_i}{dt} = A_i x_i + B \cdot n + C \cdot \Delta x_{\text{sat}}
\end{equation*}
where $A_i$ captures \emph{intrinsic dynamics} (lesion-autonomous evolution), $B$ captures local environment tumour burden (organ tumour burden through \emph{satellite count coupling}), and $C$ captures \emph{environment/satellite state change} (i.e., whether surrounding lesions move similarly or not).

We demonstrate the framework on photon-counting NSCLC CT data from metastases, recovering distinct dynamical regimes: lung lesions exhibit significant satellite count coupling ($B=-0.34$, $p<0.05$) suggesting competitive dynamics, while liver lesions show synergistic satellite behaviour coupling ($C\approx+1.0$, $p<0.05$). Synthetic validation confirms parameter recovery, and cross-coupling analysis validates that our method detects non-zero coupling when present.

This work establishes inverse dynamical inference as a principled methodology for extracting interpretable parameters from longitudinal imaging, moving beyond static feature extraction toward mechanistic characterisation of lesion behaviour.
The code and data are available at:\\
https://github.com/lukasf98/inverse-bayesian-inference

\keywords{Inverse problems \and Bayesian inference \and Longitudinal imaging \and Dynamical systems \and Spectral CT}
\end{abstract}

\section{Introduction}

Extracting dynamical parameters from discrete temporal observations is a fundamental inverse problem. Given a time series $\{x(t_k)\}$, we seek the differential equation whose solution produces the observed trajectory. This inverse problem is well-studied in physics and systems biology, yet remains underutilised in medical imaging, where longitudinal data are typically analysed through static feature extraction at each time point.

\begin{figure}[htbp!]
\centering
\includegraphics[width=0.85\textwidth]{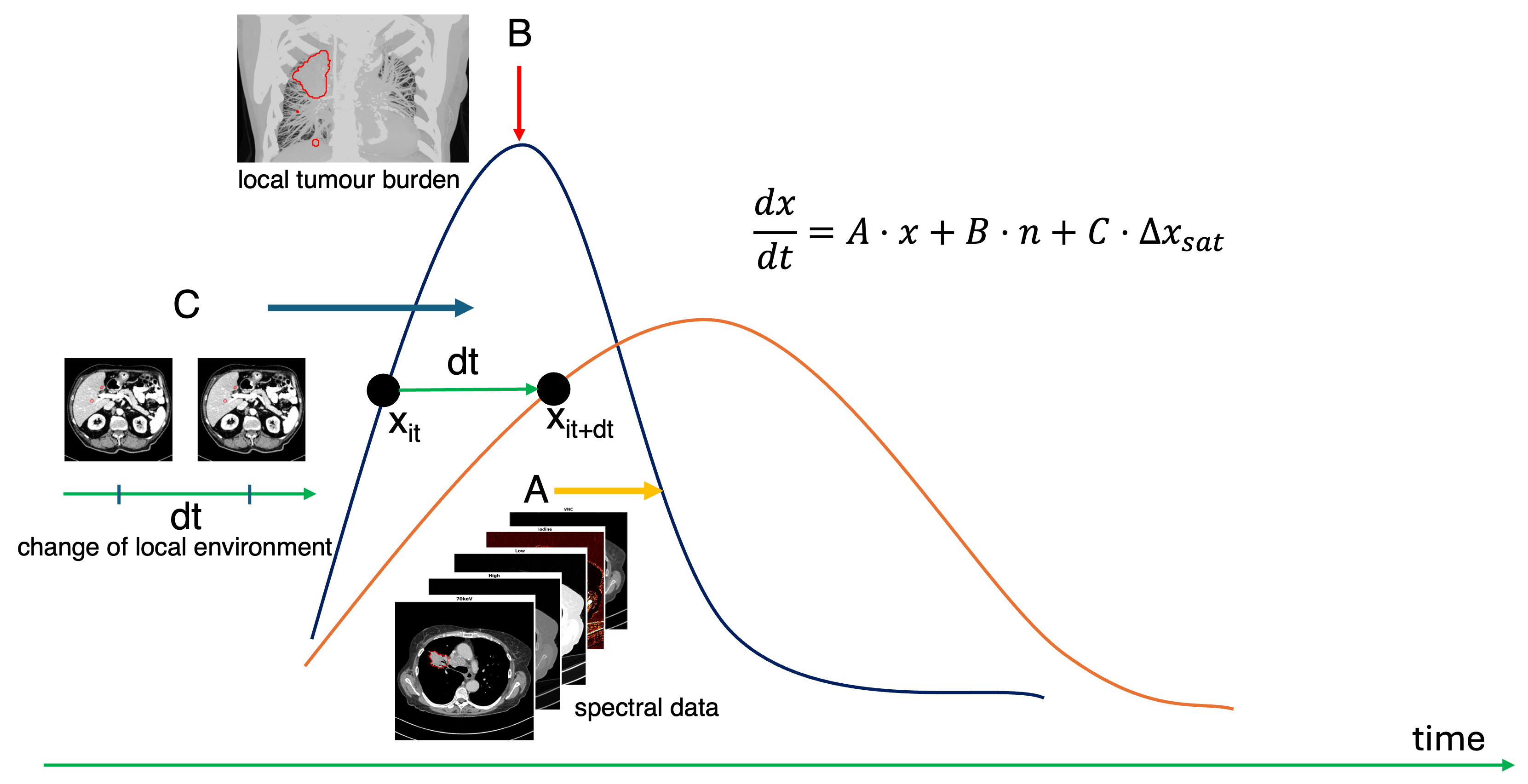}
\caption{We model the rate of change of tissue state as a linear combination of the intrinsic dynamics of the tissue (such as spectral features), the local pressure (e.g. organ tumour burden), and the change in the environment (eg. change in state of surrounding tissues). The coefficients A,B and C govern their respective contribution, and are quantifiers of the trajectory of the tissue state over time.}\label{fig:graphical_abstract}
\end{figure}

The methodological gap is significant. Standard approaches such as RECIST~\cite{eisenhauer_new_2009} reduce temporal trajectories to threshold-based classifications, while radiomic analyses~\cite{lambin_radiomics_2017} extract features independently per timepoint. Both discard the \emph{dynamical structure}, i.e., the rate parameters governing how observations evolve. These parameters encode information inaccessible from static analysis: not merely \emph{what} changed, but \emph{how fast} and \emph{in response to what}.

We address this gap by formulating inverse Bayesian inference for recovering ordinary differential equation (ODE) parameters from longitudinal imaging. The framework poses three questions about any measured quantity: (1) What is its intrinsic rate of change? (2) Does it depend on local context (e.g., number of nearby lesions)? (3) Does it couple to the behaviour of neighbouring lesions? These map to parameters $A$, $B$, and $C$ in a linear ODE, with posterior distributions providing full uncertainty quantification.

The contributions are:
\begin{enumerate}
    \item We formulate a three-component dynamical model separating intrinsic dynamics ($A$), satellite count effects ($B$), and satellite behaviour coupling ($C$), enabling systematic decomposition of lesion evolution.
    \item We develop a Bayesian inference pipeline for estimating these parameters from longitudinal imaging data with uncertainty quantification.
    \item We validate the methodology with spectral CT data, demonstrating the extraction of interpretable dynamical parameters that reveal organ-specific lesion interaction patterns.
\end{enumerate}

\section{Related Work}

Traditional approaches to longitudinal tumour analysis focus on volumetric change quantification~\cite{eisenhauer_new_2009} or radiomic feature trajectories~\cite{lambin_radiomics_2017,aerts_decoding_2014}. While radiomics extracts high-dimensional feature vectors from images, these approaches typically analyse timepoints independently or compute simple delta-features, without explicit dynamical modelling. Growth kinetics studies~\cite{mehrara_analysis_2014} fit parametric growth curves but rarely incorporate contextual factors such as satellite lesion burden.

Bayesian approaches to inverse problems provide principled uncertainty quantification~\cite{gelman_bayesian_2013,stuart_inverse_2010}. Recent work has applied Bayesian inference to biological dynamical systems~\cite{calderhead_accelerating_2008}, though applications to longitudinal imaging remain limited. Our work extends this methodology to extract lesion dynamics from serial CT data.

Photon-counting CT provides multi-energy quantitative imaging~\cite{willemink_photon-counting_2018,flohr_photon-counting_2020}, enabling material decomposition (iodine concentration, virtual non-contrast) alongside conventional attenuation at multiple energy levels. This spectral richness provides multiple channels for dynamical analysis, though temporal modelling of spectral features remains unexplored.

Our contribution bridges these areas: we apply Bayesian inverse inference to longitudinal spectral imaging, extracting interpretable dynamical parameters that characterise not only intrinsic lesion behaviour but also interactions with the local tumour microenvironment.

\section{Methods}

\subsection{Problem Formulation}

Let $x_i(t)$ denote a spectral feature (e.g., iodine concentration) of a lesion at time $t$. We seek to infer the dynamical parameters governing its evolution from discrete observations $\{x_i(t_k)\}_{k=1}^K$.

We model the rate of change as a linear combination of three components:
\begin{equation}
    \frac{dx_i}{dt} = A_i \cdot x_i + B \cdot n + C \cdot \Delta x_{\text{sat}}
    \label{eq:full_model}
\end{equation}

where:
\begin{itemize}
    \item $A_i$ are the \textbf{intrinsic dynamics} independent of context. $A_i > 0$ indicates intrinsic growth; $A_i < 0$ indicates decay.
    \item $B$ is the \textbf{satellite count coupling} effect of local tumour burden. Here $n$ is the number of lesions in the same organ. $B < 0$ suggests competition; $B > 0$ suggests facilitation.
    \item $C$ is the \textbf{environment state change} whether the lesion follows or opposes satellite lesion trajectories. Here $\Delta x_{\text{sat}}$ is the mean normalised change in spectral features across satellite lesions. $C > 0$ indicates synergistic behaviour (lesions move together); $C < 0$ indicates antagonistic behaviour.
\end{itemize}

This decomposition enables systematic hypothesis testing: Which components significantly differ from zero? Do patterns vary across anatomical contexts?
\subsection{The Inverse Problem}
Consider a dynamical system governed by an ordinary differential equation with unknown parameters $\theta$:
\begin{equation}
    \frac{dx}{dt} = f(x, \theta)
    \label{eq:forward}
\end{equation}

The \emph{forward problem} is: given $\theta$, compute the trajectory $x(t)$. The \emph{inverse problem} is: given observations $\{x(t_k)\}_{k=1}^K$, recover the parameters $\theta$ that generated them. This inverse problem is fundamental to extracting mechanistic understanding from observational data.

\subsection{Bayesian Formulation}

We adopt a Bayesian approach to the inverse problem, treating parameters as random variables and computing posterior distributions given observations. By Bayes' theorem:
\begin{equation}
    p(\theta \mid \mathcal{D}) \propto p(\mathcal{D} \mid \theta) \cdot p(\theta)
    \label{eq:bayes}
\end{equation}

where $p(\theta)$ encodes prior beliefs, $p(\mathcal{D} \mid \theta)$ is the likelihood of observing data $\mathcal{D}$ given parameters $\theta$, and $p(\theta \mid \mathcal{D})$ is the posterior distribution. This formulation provides full uncertainty quantification over inferred parameters.

\subsection{Bayesian Inference}

Given discrete observations, we approximate the derivative via finite differences:
\begin{equation}
    \frac{dx_i}{dt} \approx \frac{x_i(t_{k+1}) - x_i(t_k)}{\Delta t}
\end{equation}

For numerical stability with variable inter-scan intervals, we normalise features to baseline and set $\Delta t = 1$ (measuring rates per inter-scan interval).

The likelihood assumes Gaussian observation noise:
\begin{equation}
    p\left(\frac{dx_i}{dt} \mid A_i, B, C, \sigma\right) = \mathcal{N}\left(A_i x_i + B n + C \Delta x_{\text{sat}}, \sigma^2\right)
\end{equation}

We place weakly informative priors: $A_i, B, C \sim \mathcal{N}(0, 1)$ and $\sigma \sim \text{HalfNormal}(0.5)$. Posterior inference is performed via Hamiltonian Monte Carlo~\cite{betancourt_conceptual_2018} using PyMC, with 2000 samples after 1000 tuning iterations.

\subsection{Model Variants}

The full model (Eq.~\ref{eq:full_model}) can be reduced to test specific hypotheses:
\begin{itemize}
    \item \textbf{Intrinsic-only}: $dx_i/dt = A_i x_i$ (autonomous dynamics)
    \item \textbf{Count model}: $dx_i/dt = A_i x_i + B n$ (burden effects)
    \item \textbf{Behaviour model}: $dx_i/dt = A_i x_i + C \Delta x_{\text{sat}}$ (coordination)
\end{itemize}

We fit each variant and assess parameter significance via 95\% credible intervals.

\subsection{Predictor Orthogonality}
To assess whether $B$ (count coupling) and $C$ (behaviour coupling) can be estimated independently, we perform eigenvalue decomposition on the predictor covariance matrix $\Sigma = \text{Cov}(x, n, \Delta x_{\text{sat}})$. If the predictors are orthogonal, the eigenvalues $\lambda_1 \approx \lambda_2 \approx \lambda_3$ and the condition number $\kappa = \lambda_{\max}/\lambda_{\min} \approx 1$; the coupling terms then contribute independently to the likelihood and can be estimated in separate regressions without loss of generality. If predictors are collinear ($\kappa \gg 1$), effects may be confounded and require joint estimation or careful interpretation.

For lung lesions, eigenvalue analysis yields $\kappa < 3$ across all spectral features, confirming that $n$ and $\Delta x_{\text{sat}}$ span orthogonal directions in predictor space. The count coupling $B$ and behaviour coupling $C$ are therefore independent effects, and their separate estimation is statistically valid. For liver lesions, $\kappa > 10$ due to strong negative correlation between $n$ and $\Delta x_{\text{sat}}$ ($r < -0.6$), indicating that count and behaviour effects are entangled; liver results should be interpreted with this collinearity in mind.

\section{Experiments}

\subsection{Data}

\begin{figure}[htbp!]
\centering
\includegraphics[width=\textwidth]{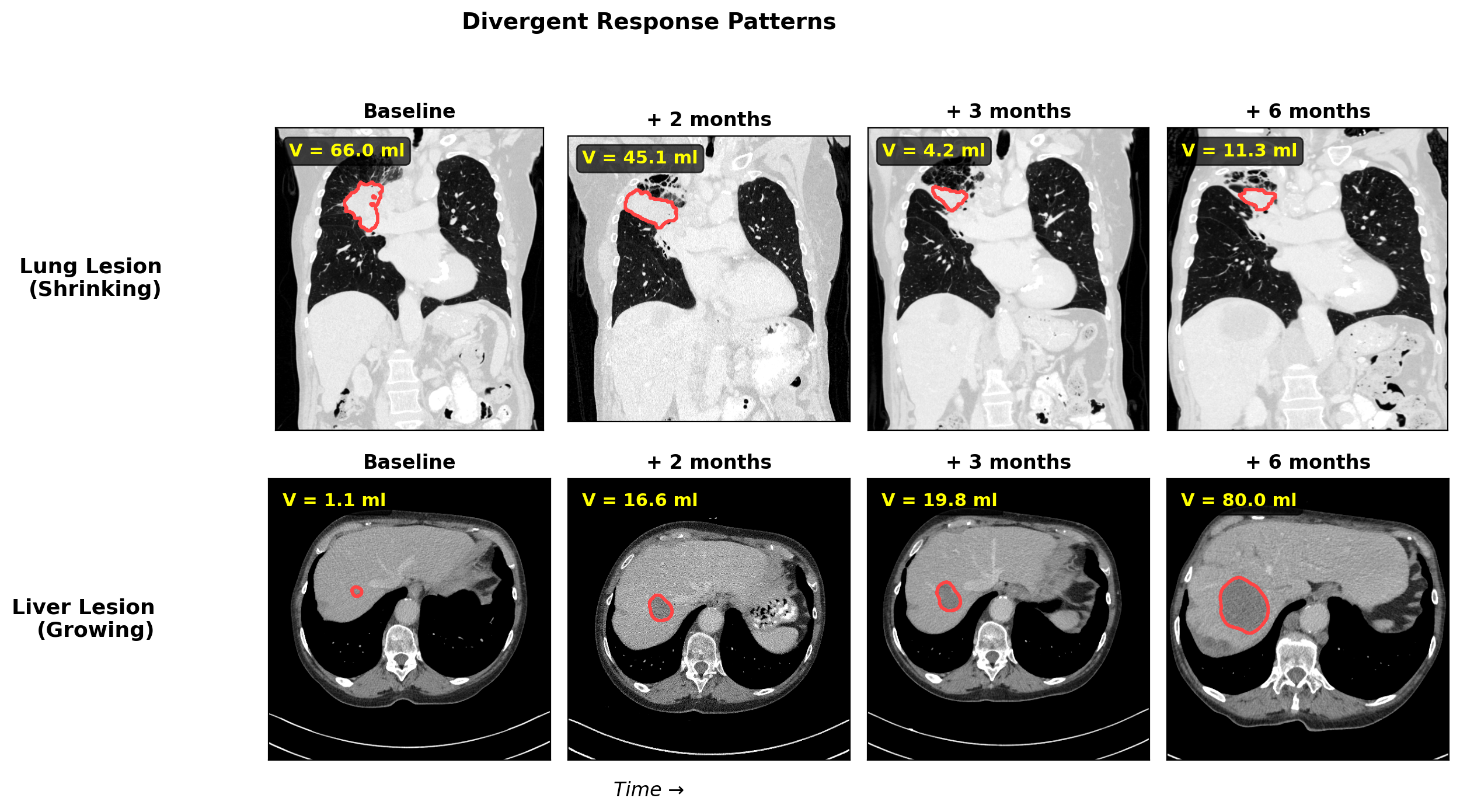}
\caption{Shrinking lung lesion and growing liver lesion for the same patient.}\label{fig:lesion_heterogenity_1}
\end{figure}

We demonstrate the framework on longitudinal photon-counting CT (NAEOTOM Alpha, Siemens Healthineers, Forchheim, DE) from NSCLC patients enrolled in an ethics approved clinical study \cite{sommer2024concept}. Per scan, we extract five spectral features: iodine concentration, virtual non-contrast (VNC), 70~keV attenuation, and high/low energy-bin values. A medical image expert tracked lesions across 2-5 time points.

\textbf{Cohort}: 32 lesions from 10 patients; 21 lung lesions (51 transitions), 11 liver lesions (19 transitions). Transitions are pairs of consecutive time points.

\subsection{Validation: Synthetic Parameter Recovery}

We generated synthetic trajectories with known parameters ($A=0.3$, $B=0.4$) for 30 lesions over 4 timepoints with observation noise ($\sigma=0.1$). The inference pipeline recovered: $\hat{A} = 0.28 \pm 0.05$ (relative error: 7\%), $\hat{B} = 0.32 \pm 0.03$ (relative error: 20\%).
Both parameters are clearly non-zero with tight credible intervals, confirming the method recovers known coupling.

\subsection{Application: Satellite Count Coupling}

We fit the count model $dx_i/dt = A_i x_i + B n$ separately for lung and liver lesions. Table~\ref{tab:count_model} presents results.

\begin{table}[t]
\caption{Satellite count coupling model. Asterisk (*) indicates 95\% CI excludes zero.}\label{tab:count_model}
\centering
\small
\begin{tabular}{llcccc}
\toprule
Organ & Feature & $A$ (intrinsic) & $B$ (count) & $B$ 95\% CI & Significant \\
\midrule
\multirow{5}{*}{Lung}
& Iodine & $+0.56$* & $-0.09$ & $[-0.32, +0.13]$ & $A$ only \\
& VNC & $+0.16$ & $-0.09$ & $[-0.40, +0.23]$ & No \\
& 70keV & $+0.25$ & $-0.07$ & $[-0.20, +0.07]$ & No \\
& High & $+0.98$* & $-0.34$* & $[-0.50, -0.18]$ & \textbf{Both}* \\
& Low & $+0.20$ & $-0.08$ & $[-0.28, +0.11]$ & No \\
\midrule
\multirow{5}{*}{Liver}
& Iodine & $+0.18$ & $-0.19$ & $[-0.73, +0.36]$ & No \\
& VNC & $+0.08$ & $-0.04$ & $[-0.22, +0.13]$ & No \\
& 70keV & $+0.25$ & $-0.13$ & $[-0.38, +0.11]$ & No \\
& High & $+0.20$ & $-0.12$ & $[-0.34, +0.10]$ & No \\
& Low & $+0.31$ & $-0.16$ & $[-0.40, +0.08]$ & No \\
\bottomrule
\end{tabular}
\end{table}

For \textbf{lung lesions}, the High-keV channel shows significant satellite count coupling: $B = -0.34$ (95\% CI: $[-0.50, -0.18]$). The negative sign indicates \emph{competitive dynamics}: more satellite lesions in the lung correlate with reduced High-keV signal evolution. This suggests resource competition among co-located lesions. High-keV signal informs of attenuation in high energy levels, and thus, it is a surrogate for heavy element concentration. In practical terms, this is explained by increased blood supply (i.e., iodine concentration), but possibly, storage of other heavier elements such as calcium in typical calcifications seen in tumours. Changes in the heavy element concentration in lung can be picked far easier in CT due to the base low density of lung tissue.

For \textbf{liver lesions}, no significant count coupling was detected, though the cohort is smaller (19 vs 51 observations).

\subsection{Application: Satellite Behaviour Coupling}

We fit the behaviour model $dx_i/dt = A_i x_i + C \Delta x_{\text{sat}}$ to test whether lesions move synergistically or antagonistically with their satellites. Table~\ref{tab:behaviour_model} presents results.

\begin{table}[t]
\caption{Satellite behaviour coupling model. $C > 0$: synergistic; $C < 0$: antagonistic.}\label{tab:behaviour_model}
\centering
\small
\begin{tabular}{llcccc}
\toprule
Organ & Feature & $A$ (intrinsic) & $C$ (behaviour) & $C$ 95\% CI & Pattern \\
\midrule
\multirow{5}{*}{Lung}
& Iodine & $+0.55$* & $-0.00$ & $[-0.22, +0.22]$ & Independent \\
& VNC & $+0.08$ & $-0.01$ & $[-0.06, +0.04]$ & Independent \\
& 70keV & $+0.09$ & $+0.17$ & $[-0.17, +0.53]$ & Independent \\
& High & $+0.65$* & $+0.13$ & $[-0.12, +0.38]$ & Independent \\
& Low & $-0.08$ & $+0.03$ & $[-0.26, +0.31]$ & Independent \\
\midrule
\multirow{5}{*}{Liver}
& Iodine & $-0.92$* & $+0.80$ & $[-0.50, +2.03]$ & Independent \\
& VNC & $+0.00$ & $+0.70$* & $[+0.03, +1.25]$ & \textbf{Synergistic}* \\
& 70keV & $-0.05$ & $+0.95$* & $[+0.19, +1.62]$ & \textbf{Synergistic}* \\
& High & $-0.03$ & $+1.01$* & $[+0.56, +1.39]$ & \textbf{Synergistic}* \\
& Low & $-0.09$ & $+0.85$ & $[-0.19, +1.73]$ & Independent \\
\bottomrule
\end{tabular}
\end{table}

For \textbf{liver lesions}, strong synergistic coupling emerges: $C \approx +1.0$ for VNC, 70keV, and High-keV (all $p < 0.05$). This indicates that liver lesions move together: when satellite lesions increase in density, the primary lesion does too. The coupling strength near unity suggests near-perfect coordination. Biologically, VNC is a surrogate for tissue density, while 70keV combines blood supply with density. A synergistic effect is plausible considering that tumor-induced arterial angiogenesis in the liver builds on increased demand for the body to increase arterial vascularity in an area mainly supplied venously.

For \textbf{lung lesions}, no significant behaviour coupling was detected. Combined with the count coupling result, this suggests lung lesions are affected by the \emph{number} of satellites (competition) but not their \emph{behaviour}.

\subsection{Cross-Organ Effects}

We tested whether satellite burden in one organ affects lesion dynamics in another (systemic effects). Using the model $dx_i/dt = A_i x_i + B_{\text{remote}} n_{\text{other}}$:
\begin{itemize}
    \item Lung lesion count $\to$ Liver dynamics: $B \approx 0$ (not significant)
    \item Liver lesion count $\to$ Lung dynamics: $B \approx 0$ (not significant)
\end{itemize}

Satellite effects are \emph{organ-specific}, not systemic. Local microenvironment governs lesion interactions.

\section{Discussion}

We have presented inverse Bayesian inference as a framework for extracting dynamical parameters from longitudinal imaging. The key methodological contributions are:

\begin{enumerate}
    \item \textbf{Explicit dynamical modelling}: Rather than treating timepoints independently, we model the generative process underlying temporal evolution.
    \item \textbf{Decomposition into interpretable components}: The $A$-$B$-$C$ formulation separates intrinsic dynamics from contextual effects, enabling systematic hypothesis testing.
    \item \textbf{Uncertainty quantification}: Bayesian inference provides credible intervals, distinguishing significant effects from noise.
    \item \textbf{Validation framework}: Synthetic recovery and cross-coupling analysis verify that the method detects coupling when present.
\end{enumerate}

Application to spectral CT data revealed organ-specific interaction patterns. For lung lesions, where eigenvalue analysis confirmed predictor orthogonality ($\kappa < 3$), we can cleanly separate effects:

\begin{itemize}
    \item \textbf{Count coupling} ($B = -0.34$, significant): More satellites correlate with reduced High-keV evolution, suggesting competitive dynamics.
    \item \textbf{Behaviour coupling} ($C \approx 0$, not significant): Satellite trajectories do not predict primary lesion trajectories.
\end{itemize}

The dissociation that lesions compete for resources ($B < 0$) while evolving independently ($C \approx 0$), would not be discoverable without the decomposed framework and orthogonality verification.

Notably, the intrinsic term for iodine is consistently positive and significant ($A \approx +0.55$, $p < 0.05$) across both coupling analyses for lung metastases, indicating that iodine concentration increases autonomously regardless of satellite count or behaviour. This suggests iodine uptake follows an intrinsic trajectory decoupled from local tumour burden effects, matching the expected strong tumour-triggered angiogenesis of NSCLC lung metastases \cite{hall_angiogenesis_2015}.

For liver lesions, predictor collinearity ($\kappa > 10$) prevents clean separation of $B$ and $C$. The nominally significant $C \approx +1$ may reflect entanglement with count effects rather than true behavioural synchrony.

These patterns were not assumed \emph{a priori} but emerged from the inference procedure, demonstrating the framework's ability to discover structure in longitudinal data.

The limitations of the method stem from not capturing the nonlinear dynamics with the defined ODE system. The cohort size limits statistical power, particularly for liver lesions. Future work should explore nonlinear models, larger cohorts, and extension to other imaging modalities such as PET where modelling of tracer uptake is highly informative.

We have established inverse Bayesian inference on ODEs as a methodology for extracting lesion dynamics from longitudinal imaging. The framework decomposes temporal evolution into intrinsic, count-dependent, and behaviour-dependent components, enabling principled hypothesis testing about lesion interactions. Application to spectral CT revealed plausible organ-specific patterns, specifically competitive dynamics in lung, and synergistic dynamics in liver. These findings demonstrate that the approach extracts interpretable, context-dependent dynamical parameters from imaging data.

\begin{credits}
\subsubsection{\ackname} This research was partially funded by the Intramural Research Funding Grants ``AI-driven Longitudinal Lesion Tracking'' of the Faculty of Medicine, University of Augsburg, the Bavarian Center for Cancer Research as part of the Lighthouse ``Local Therapies'' and the Study Group ``Surrogate parameters for CNS Tumors in Childhood'', as well as by the Bavarian Ministry of Economic Affairs, Regional Development and Energy (StMWi) under grant number DIK-2310-0004//DIK0556/02.
\subsubsection{Disclosures.}
The authors used Claude Opus 4.8 (Anthropic) to assist with drafting and editing the text of this manuscript. All AI-assisted content was reviewed, verified, and edited by the authors.
\end{credits}

\bibliographystyle{splncs04}
\bibliography{NSCLC}

\end{document}